\documentclass{AUJarticle}
\usepackage[cmex10]{amsmath}
\usepackage[utf8x]{inputenc}
\usepackage[nocompress]{cite}
\usepackage{graphicx, multirow, booktabs, color, listings}
\usepackage{balance}

\usepackage{subcaption}
\usepackage{caption}
\usepackage{array}
\usepackage{multirow}
\usepackage{cleveref}
\usepackage{xurl}

\graphicspath{{figures/}}

\begin{document}

\title{Denoising Autoencoder-based Defensive Distillation as an Adversarial Robustness Algorithm}

\addauthor{Bakary Badjie, José Cecílio, António Casimiro}
{LASIGE, Departamento de Informática, Faculdade de Ciências da Universidade Lisboa, Lisboa} 
  {\{bbadjie, jmcecilio, casim\}@ciencias.ulisboa.pt}

\issuev{1}
\issuen{1}
\issued{March 2023}

\shortauthor{B. Badjie et al.}
\shorttitle{Denoising Autoencoder-based Defensive Distillation}

\thispagestyle{plain}

\maketitle

\begin{abstract}


Adversarial attacks significantly threaten the robustness of deep neural networks (DNNs). Despite the multiple defensive methods employed, they are nevertheless vulnerable to poison attacks, where attackers meddle with the initial training data. In order to defend DNNs against such adversarial attacks, this work proposes a novel method that combines the defensive distillation mechanism with a denoising autoencoder (DAE). This technique tries to lower the sensitivity of the distilled model to poison attacks by spotting and reconstructing poisonous adversarial inputs in the training data. We added carefully created adversarial samples to the initial training data to assess the proposed method's performance. Our experimental findings demonstrate that our method successfully identified and reconstructed the poisonous inputs while also considering enhancing the DNN's resilience. The proposed approach provides a potent and robust defence mechanism for DNNs in various applications where data poisoning attacks are a concern. Thus, the defensive distillation technique's limitation posed by poisonous adversarial attacks is overcome.

Keywords: \textit{Deep Neural Network}, \textit{Denoising Autoencoder}, \textit{Defensive Distillatiomn}, \textit{Adversarial attacks and Robustmness}, \textit{Alibi-detect}

\end{abstract}


\section{Introduction}
The necessity of DNNs' resilience against adversarial attacks has shown a growing concern as their use in real-world safety-critical systems has increased exponentially. In this context, adversarial attacks \cite{chen2022adversarial} attempt to trick a DNN by making subtle alterations to the input data that are imperceptible to humans but have a significant negative impact on the DNN's ability to make accurate decisions. Several defence strategies have been proposed to overcome this problem, including defensive distillation, which has been successful in defending DNNs from adversarial attacks in run-time settings \cite{papernot2016distillation}. Nevertheless, one of the drawbacks of the defensive distillation method is that it remains susceptible to data poisoning attacks, in which adversaries aim to impair the model's performance by inserting erroneous data entries into the training set. These harmful data points could be carefully crafted to be close to the model's decision boundary to circumvent the countermeasures offered by defensive distillation. This study presents a novel method for robustifying a distilled network against data poisoning adversarial threats by integrating a denoising autoencoder (DAE) \cite{bank2020autoencoders} in the defensive distillation cycle. Defensive distillation involves training two DNNs, the instructor model (the first model), and distilling its knowledge into the student model (the second or distilled model) to make it robust adversarial examples and previously unseen input \cite{chen2022adversarial}. A DAE is a type of DNN that is trained to detect and discard noise from input data and reconstruct it back to its original form \cite{bank2020autoencoders}.

This paper is motivated by the fact that the instructor model is not immune to data poisoning adversarial attacks. Although the student model has more latitude to reject input modifications because it leverages the "distilled" version of the training data, where the training examples are transformed by a temperature parameter $T$ ~\cite{papernot2016distillation}. 
Thus, minimising the instructor model's susceptibility to data poisoning attacks is pivotal for developing a reliable, distilled DNN. To achieve this, we designed a DAE to detect and reconstruct poisonous adversarial inputs in the training data. The defensive distillation method already offers a strong foundation, but when combined with a DAE, the protection mechanism against data poisoning adversarial attacks is significantly strengthened. Moreover, our strategy considers several adversarial attack aspects, such as the attacker's access to the training data and trial-and-error techniques to access the model gradient, making it a more effective defence mechanism against such attacks.

To evaluate the effectiveness of the proposed algorithm, we used the fast gradient sign method (FGSM) \cite{goodfellow2014explaining} and the iterative-fast gradient sign method (I-FGSM) \cite{kurakin2016adversarial}, which uses insusceptible perturbations, to modify some of the initial training inputs to craft poisonous samples. 

The results show that the approach could detect and reconstruct the poisonous inputs in the training data and significantly improve the robustness of the distilled DNN against adversarial poisoning attacks. Although the alteration made by FGSM and I-FGSM on the training data is so subtle that humans are unlikely to notice them, the DAE detected them in the training data and reconstructed them back to their original form before feeding them into the instructor network. This made the distilled model less vulnerable to attack at runtime. The proposed approach offers a more potent protection mechanism against adversarial poisoning attacks because it combines defensive distillation with a DAE. The limitation of the defensive distillation approach brought on by data poison attacks has been overcome by this approach. 

Statistically and experimentally, Papernot et al. \cite{papernot2016distillation} explore distillation as a preventive measure against adversarial inputs. Using information distilled from the instructor network, the authors minimize the amplitude of student network gradients that attackers might access to generate adversarial examples. Moreover, they demonstrate that models developed via defensive distillation are less susceptible to antagonistic inputs in runtime settings. Additionally, according to Goldblum, Micah, et al. \cite{goldblum2020adversarially}, while effective neural networks (NN) may be produced by transferring information from the teacher model to the student model, they may still be exposed to strong adversarial onslaughts in run-time scenarios. Adversarially Robust Distillation (ARD), which creates tiny models that distil robustness in larger networks, was proposed by the same authors as a solution to this problem. The authors say this strategy performs better than the conventional defensive distillation benchmark. Previous works where the defensive distillation technique was leveraged include \cite{kuzlu2022adversarial, chen2021ltd, papernot2017extending}. Although these strategies are successful in preventing evasion attacks (run-time attacks), none of them has taken poison adversarial attacks into consideration. To address this shortcoming, this paper integrated a DAE in the knowledge distillation process to serve as a filter during the training phase to protect against data poisoning attacks.  Yue et al. \cite{yue2021denoising} demonstrated that DAEs are highly helpful in spotting and reconstructing contaminated images. In their article, they suggest a DAE-based approach for identifying and mitigating poison adversarial attacks in federated learning. This involves training machine learning (ML) models on distributed data from many sources. Other works where DAE is used as a filter against poison data include \cite{kascenas2022denoising, tripathi2021facial}.

\section{Mitigation of adversarial poison attacks}
Defensive distillation has been widely used to improve the resilience of ML models against adversarial attacks. It has shown impressive success but is still vulnerable to data poisoning attacks, in which the adversaries alter the initial training data. This work incorporates DAE into the knowledge distillation training process to defend against such onslaughts.


The technique used in defensive distillation involves training two DNN models that are similar in structure (i.e. the instructor and the student models). The instructor model $f_1$ is trained using the original dataset, and it learns normally with the inclusion of a softmax temperature $T$, which is increased to $5$ degrees in our case. This enables the softmax layer of $f_1$ to produce a very soft probability vector $F(X)$, closer to a uniform statistical distribution of the data. This $F(X)$ is used to label each training data point to train the second model $f_2$, which is the student model. $T$ is usually reset to its default value of $1$ during testing. This method multiplies the features that must be altered in the $f_2$ to create a successful attack. Attackers would need to modify many features in the input to drive $f_2$ into incorrect predictions, making it considerably more challenging to construct adversarial inputs. Also, it lessens the model's gradient, which an attacker may use to breach the model.

\subsection{Denoising Autoencoder}

The goal of the DAE is to learn a compressed representation of the input data, correct any abnormalities in the data, and reconstruct it back to its original, undistorted state with the help of the latent vector $h$. The design comprises an encoder network $f(x^*)$ and a decoder network $f(x)$, with $f(x^*)$ receiving the erroneous form of data $x^*$ and translating it to a lower-dimensional latent representation $h$, which is used by $f(x)$ to reconstruct the $x^*$ to its original form. During training, the DAE learns how to minimize the reconstruction error between the initial unperturbed data input $X$ and output reconstructed data $x$. This is accomplished by learning the mapping function that is resilient to noise or adversarial perturbation $\epsilon$ and its ability to rebuild the $x^*$ to their original form without compromising the most significant intrinsic statistical properties of $x$. The latent representation can be expressed as;

\begin{equation}
  h = f(Wx + b)   
\end{equation}

where:
$h$ is the latent representation vector
$x$ is the input data vector
$W$ is the weight matrix
$b$ is the bias vector
$f$ is the activation function applied element-wise to the linear transformation of the input data, which introduces non-linearity to the model.

\subsection{Adversarial Attacks and Robustness}

Adversarial attacks are types of hostile modification of the data that seek to deceive ML models in their decision-making \cite{chen2022adversarial}. A wide variety of ML applications are susceptible to adversarial attacks. In contrast, adversarial robustness describes the model's capacity to maintain its expected performance in the face of interruptions or hostile attacks \cite{deng2020analysis}. A model is deemed resilient if it can correctly classify input data, even in the presence of minor alterations or malicious attacks. Robustness is a crucial characteristic for ML models since it guarantees their dependability in real-world applications where the input data may be noisy or purposefully altered \cite{wang2022deep}. In our experiment, we used the following parameters to generate adversarial perturbations for FGSM and IFGSM; $epsilon_fgsm = 0.01, epsilon_ifgsm = 0.01, alpha = 0.01, num_iterations = 10$. The epsilon $\epsilon$ is a parameter that detects the magnitude of the perturbation, which should be used to alter the image's pixel values. In contrast, the alpha $\alpha$ parameter controls the step size of the perturbations in each iteration of the IFGSM attack. $num_iterations = 10$ indicates that the IFGSM will be applied at every $10$ iteration.

\section{Autoencoder-based Defensive Distillation Approach}

Following the methodology used \cite{papernot2017extending}, the student model in our study is built to be uncertain about its prediction when the input required to classify is statistically far from the training set. However, we used the Kullback–Leibler (KL) divergence \cite{raiber2017kullback} to quantify the statistical differences between the input and the training data. If their statistical difference is more significant than the $P-value$, the network becomes uncertain about its prediction on that data point and will return a null value. This ambiguity is evaluated through a technique termed "dropout inference," in which a proportion of neurons in a NN are arbitrarily eliminated during inference. This approach allowed the model to offer uncertainty estimates for the predictions and estimate Bayesian inference. Analyzing the teacher's model's predictions yields the uncertainty measurements needed to train the student model. The soft probability vector $F(X)$ produced by the teacher model through its softmax function with $T=5$ is not enough to harden the student model against adversarial perturbation as done in \cite{papernot2016distillation} because this will make the student model classify inputs based on the softmax traditional probability estimation without considering the uncertainty of its prediction. The output probability vector $F(X)$ of the softmax function layer is computed as:

\begin{equation} \label{eq-1}
   F(X) =  \left[\frac{e^{Z_i\frac{(X)}{T}}}    {\sum_{l=0}^{N-1} e^{z_l\frac{(X)}{T}}} \right]_{i \in 0----N-1} 
\end{equation}

A given neuron within the softmax layer that corresponds to a class indexed by $i \in 0..N -1$,  where N is the number of classes computes component $i$, $X$ is the data to each input, and $z$ is the output of the last hidden layer of the network.

While developing the DAE model, we used the "Keras tuner library", a TensorFlow-based hyperparameter optimization toolkit, to select the appropriate hyperparameters. We leveraged FGSM and IFGSM attack algorithms to generate adversarial inputs from a portion of the Germany Traffic Sign Recognition Benchmark (GTSRB) dataset while the remaining portion is used as the clean data. Mean squared error (MSE) loss is used to evaluate the reconstruction error between $X$ and $x$, where the goal is to minimize the statistical difference between the $x^*$ and $x$. Furthermore, to avoid issues with vanishing or exploding gradients and promote quicker and more efficient convergence during training, we use the random weight initialization strategy. In the training phase, both $X$ and $x^*$ are fed to the DAE as inputs to enable it to learn the statistical correlations between them. During backpropagation, the loss function is minimized while the weights are updated using a stochastic gradient descent (SGD) optimization algorithm. The effectiveness of the trained DAE is evaluated on a different test dataset containing both clean and distorted images. A specified reconstruction threshold is set, which serves as a yardstick to clarify adversarial examples after reconstruction. The images whose reconstruction error is above this threshold are termed adversarial inputs, and they are subsequently get discarded before reaching the teacher model. 

\section{Results}

\begin{figure}[!tb] 
 \centering
 \includegraphics[width=1.0\columnwidth]{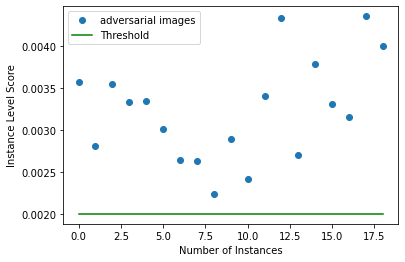} 
 \caption{Illustration of the DAE's performance on only adversarial inputs during reconstruction} 
 \label{fig-adversarial}
\end{figure}

\begin{figure}[!tb] 
 \centering
 \includegraphics[width=1.05\columnwidth]{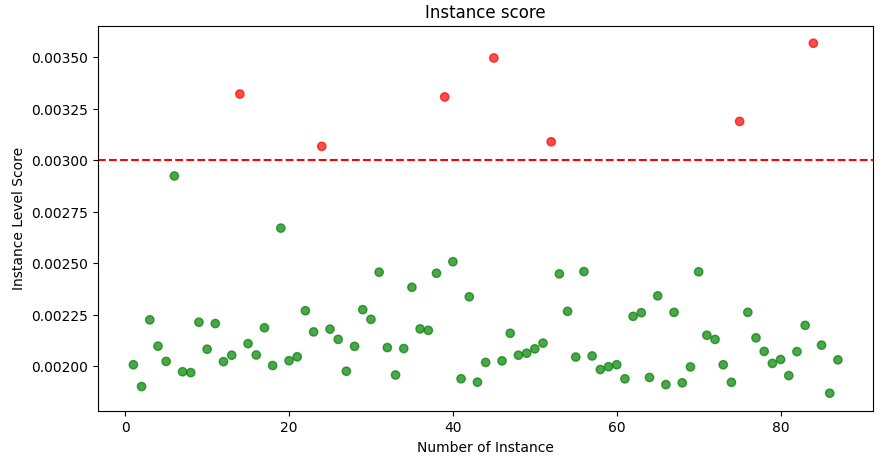} 
 \caption{Illustration of the DAE's performance on the evaluation dataset during reconstruction.}
  \label{fig-clean_and_bad}
\end{figure}

A starting reconstruction threshold of $0.015$ was chosen randomly at the beginning of the DAE network's training process. This made it easier to update the reconstruction threshold to $0.003$ during the evaluation phase that followed. We used the $"infer threshold"$ function to update the threshold, which is part of the $"alibi-detect"$ outlier/adversarial detection library. The "infer threshold" function calculates the values inferred from the percentage of instances identified as adversarial in the test or evaluation dataset. Figures \ref{fig-adversarial} and \ref{fig-clean_and_bad} show how the DAE network performed on the evaluation datasets, which comprised instances of only adversarial inputs and a combination of both adversarial and clean inputs.
 
\begin{figure}[!tb] 
 \centering
 \includegraphics[width=1.05\columnwidth]{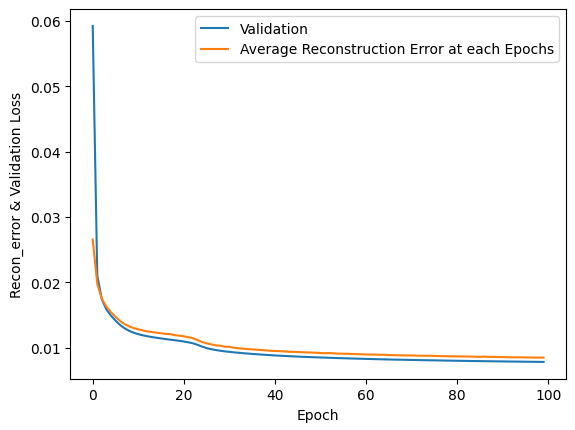} 
 \caption{Illustration of the DAE's performance on the evaluation dataset during reconstruction.}
  \label{fig-loss_error}
\end{figure}

 Upon reconstruction, images with reconstruction errors greater than the new threshold value $(0.003)$ are labelled as adversarial inputs since the DAE network is unable to restore the images to their original state. Reconstructed-clean images are those with reconstruction errors below the new threshold value. Before passing the dataset to the instructor model in the distillation stage, the adversarial inputs are subsequently eliminated. As shown in Figure \ref{fig-clean_and_bad}, the green dotted circles indicate the reconstructed clean images, whereas the red dotted circles represent the adversarial images that the DAE network was unable to reconstruct. According to our experimental results, the average reconstruction error produced by the DAE network using the evaluation dataset was $0.008190$, indicating that our designed DAE network functioned satisfactorily. The average reconstruction and validation loss at each epoch are shown in Figure \ref{fig-loss_error}.

The result also demonstrates that IFGSM devises stronger adversarial examples than FGSM because FGSM only takes into account the gradient of the loss function with respect to the input once and then makes a single step along the path of the gradient to create an adversarial example, whereas IFGSM takes into account the gradient with respect to the input at every iteration and keeps adding a small perturbation to the input in the direction of the gradient's sign, which results in stronger adversarial examples.

The reconstructed images are sent to the teacher model, which is trained with a softmax temperature of $T = 5$. This produces soft labels to annotate the new dataset, which we then used to train the student model (distilled network). The instructor model's total accuracy throughout training is $99.89\%$, with an average loss of $0.11\%$. The student model also performed well by correctly classifying adversarial inputs in the test dataset with an accuracy rate of $76.09\%$. Although this precision might not be particularly excellent in other situations, the design is incredibly robust to IFGSM and FGSM adversarial attacks.

\section{Conclusion}

The use of a DAE network in combination with defensive distillation has proven to be an effective method to robustify a distilled network against both poisoning and evasion adversarial attacks. It was also shown that the "alibi-detect" 
library helped the DAE network detect and get rid of adversarial inputs in the training dataset. By updating the reconstruction threshold during the evaluation phase, the DAE network was able to accurately identify inputs as either adversarial or clean. The use of IFGSM resulted in stronger adversarial examples than FGSM, highlighting the importance of considering the gradient of the loss function with respect to the input at every iteration. The student model, trained with soft labels produced by the teacher model, demonstrated excellent robustness against adversarial attacks. Based on the results, the proposed approach could be a valuable addition to the arsenal of techniques used to improve the security of ML systems.

\section{Acknowledgments}
This work was supported by the LASIGE Research Unit (ref. UIDB/00408/2020 and ref. UIDP/00408/2020), and by the European Union’s Horizon 2020 research and innovation programme under grant agreement No 957197 (VEDLIoT project).








\bibliographystyle{ieeetr}
\bibliography{References}
\balance

\end{document}